\documentclass{article}
\usepackage{fullpage}

\usepackage{array}
\usepackage{multirow}
\usepackage{listings}
\usepackage{xspace}
\usepackage{xcolor}
\usepackage{hyperref}
\usepackage{amsfonts}
\usepackage{amssymb}
\usepackage{amsmath}
\usepackage{graphicx}
\usepackage{booktabs}
\usepackage{placeins}

\usepackage{subcaption}
\usepackage{threeparttable} % table notes that render beneath the table (\footnote fails inside tabular)

\newcommand{\code}[1]{\texttt{#1}}
\newcommand{\nquantizers}{25\xspace}
\newcommand{\nimplemented}{14\xspace}
\newcommand{\ncatalogimplemented}{10\xspace}
\newcommand{\ncatalogsupported}{12\xspace}
\newcommand{\ncatalogunsupported}{3\xspace}

\newcommand{\nprimitives}{7\xspace}
\newcommand{\nprimitivesfunction}{37\xspace}
\newcommand{\vqb}{VQ-bench\xspace}
\makeatletter
\def\cp@stop{\cp@stop}
\def\cp@scan#1{%
  \ifx#1\cp@stop\else
    #1%
    \ifx#1.\penalty0 \fi
    \ifx#1,\penalty0 \fi
    \ifx#1[\penalty0 \fi
    \expandafter\cp@scan
  \fi}
\newcommand{\codepipe}[1]{\texttt{\cp@scan#1\cp@stop}}
\makeatother

\newcommand{\prg}[1]{%
  \par\addvspace{0.35\baselineskip}%
  \noindent\textbf{#1}\hspace{0.6em}\ignorespaces}
\newcommand{\R}{\mathbb{R}}
\newcommand{\E}{\mathbb{E}}

\newcommand{\resultgrid}[4][legend-2col]{%
  \begin{figure}[ht]
    \centering
    \begin{subfigure}[t]{0.245\columnwidth}%
      \centering\includegraphics[width=\linewidth]{attachments/recon-mse-#2}%
      \caption{Reconstruction MSE}%
    \end{subfigure}\hfill
    \begin{subfigure}[t]{0.245\columnwidth}%
      \centering\includegraphics[width=\linewidth]{attachments/score-mse-#2}%
      \caption{Score MSE}%
    \end{subfigure}\hfill
    \begin{subfigure}[t]{0.245\columnwidth}%
      \centering\includegraphics[width=\linewidth]{attachments/recall10-#2}%
      \caption{Recall@$10$}%
    \end{subfigure}\hfill
    \begin{subfigure}[t]{0.245\columnwidth}%
      \centering\includegraphics[width=\linewidth]{attachments/attention-tv-#2}%
      \caption{Attention TV ($\tau=0.05$)}%
    \end{subfigure}\\[6pt]
    \includegraphics[width=0.50\columnwidth]{attachments/#1}%
    \caption{#3}%
    \label{#4}%
  \end{figure}%
}

\title{VQ-bench: A Composable Vector Quantization Framework}

\author{
  Ashwin Padaki\footnote{University of Pennsylvania and Pinecone, \texttt{apadaki@seas.upenn.edu}} \and
  Amir Ingber\footnote{Pinecone, \texttt{ingber@pinecone.io}}  \and
  Edo Liberty\footnote{Pinecone, \texttt{edo@pinecone.io}}
}

\begin{document}
\maketitle

%\keywords{\href{www.vq-bench.com}{www.vq-bench.com}, VQ-bench, vector quantization, benchmarking, approximate nearest neighbor search, model compression, KV cache, reproducibility}

\begin{abstract}
Vector quantization is an old problem but has recently become central to AI infrastructure. 
It is therefore experiencing a surge of renewed engineering and research activity. 
This paper provides a unified framework for developing and benchmarking new quantization algorithms. 
We describe \nprimitives common conceptual quantization \emph{primitives} and show how to compose them arbitrarily. 
We then re-express \nquantizers common quantizers as \emph{pipelines} of these primitives. 
Finally, we publish \vqb as open-source to be extended further and make reproducible benchmarks publicly available.
\end{abstract}

% =====================================================================
\section{Introduction}
\label{sec:intro}

Vector quantization (VQ) is the following problem. Given a matrix $X\in\R^{d\times n}$ of $n$ vectors in dimension $d$, and optionally a matrix $Q'\in\R^{d\times m}$ of $m$ sample queries, produce a $b$-bit \emph{code} for each vector and a \emph{model}. Then, using just the codes and the model, support one of two tasks: (1) approximately \emph{reconstruct} the original vectors, computing $\tilde{X} \approx X$, or (2) given new queries $Q$, approximate their inner product \emph{scores} with $X$, computing $\tilde{S}\approx Q^T X$.

Most of the core ideas and techniques for this problem were already known by the  mid-1980s; 
the seminal book by Gersho and Gray was published in 1991 and recounts the history of VQ~\cite{Gersho1991VectorQA}; 
see \S\ref{sec:review} for more historical context. 
Nevertheless, quantization is, again, at the center of attention because it became crucial to modern AI. 
Vector databases like \href{https://www.pinecone.io}{Pinecone} use it for large scale approximate nearest neighbor (ANN) search~\cite{jegou2011pq,johnson2017faiss,simhadri2022bigann,aumuller2026results} 
which powers retrieval-augmented generation (RAG) and agentic search. 
LLM serving relies on quantizing model weights to reduce their memory footprint~\cite{frantar2023optq,lin2023awq,dettmers2023qlora}. 
Finally, transformer-based models save their context in vector-valued key-value (KV) caches. 
As context windows grow, the KV-cache size becomes the bottleneck for GPU memory capacity and bandwidth~\cite{liu2024kivi,hooper2024kvquant,kang2024gear}.

Progress, however, came with a price. Keeping track of the state of the art
and assessing the effectiveness and novelty of new results has become
increasingly difficult. Many tens of papers were published on quantization in the last few
years alone, and with AI-aided experimentation and paper authoring, the rate
is only increasing. These results are often difficult to verify
independently, let alone compare to one another. Each
publication reports results on different datasets, measures different quality
metrics, runs on different hardware, and accounts for resource consumption differently. 
Variations in engineering investment further blur 
the line between improved implementation quality and genuine algorithmic improvements.
Finally, credit attribution across the field is usually incomplete, which
can make it even harder to distinguish genuinely novel ideas from rediscoveries or variations of existing algorithms.

% A recent example is TurboQuant~\cite{zandieh2025turboquant}, whose core components 
% (a random rotation followed by scalar quantization) overlap with the earlier
% RaBitQ~\cite{gao2024rabitq,gao2026revisiting} and
% DRIVE/EDEN line~\cite{vargaftik2022eden,turboquant_eden_note}.

\subsection{Our contributions}
In this paper, we attempt to bring some order to the chaos.
We begin with describing how to build arbitrary quantizers from a short list of algorithmic primitives.

\emph{Primitives} perform a specific operation on vectors, queries, and
scores (e.g.,\ centering, rotating, splitting coordinates, rounding to integers, etc.).
We list and explain all core primitives in 
Section~\S\ref{sec:primitives} and summarize them in Table~\ref{tab:primitives}.
We then provide a unified compositional interface so they can be chained into complete quantizers. 

\emph{Quantizers} are created by pipelines of primitives. 
For example, 2-bit-per-dimension product quantization~\cite{jegou2011pq} can be concisely expressed as
\code{split(segment,width=4).}\allowbreak\code{kmeans(k=256)}.
We survey \nquantizers\ published quantizers as such pipelines (\S\ref{sec:quantizers},
Table~\ref{tab:catalog}). 

Since our framework allows for arbitrary composition, one can easily experiment with variations of existing quantizers or invent new ones. 
To facilitate that, we open-source an implementation of this framework. 
\vqb provides Rust implementations of these primitive functions.
It currently runs \nimplemented\ quantizers end-to-end: 12 published methods and two baselines.
We also publish results at \href{https://vq-bench.com}{vq-bench.com} for these quantizers and measure 9 quality metrics.
The datasets we use for benchmarking are in the common hdf5 format (details below), and we start with 5 datasets from \cite{jaasaari2025}.
\vqb also measures compression rates, running time, and memory
consumption. \vqb aspires to follow in the footsteps of other successful benchmarking
suites that pushed the field forward: Faiss~\cite{douze2025faisslibrary},
ANN-Benchmarks~\cite{aumuller2017annbench}, 
BigANN~\cite{simhadri2022bigann,aumuller2026results}, and, more recently, VIBE~\cite{jaasaari2025}.

\begin{figure}[t]
\centering
\includegraphics[width=0.8\linewidth]{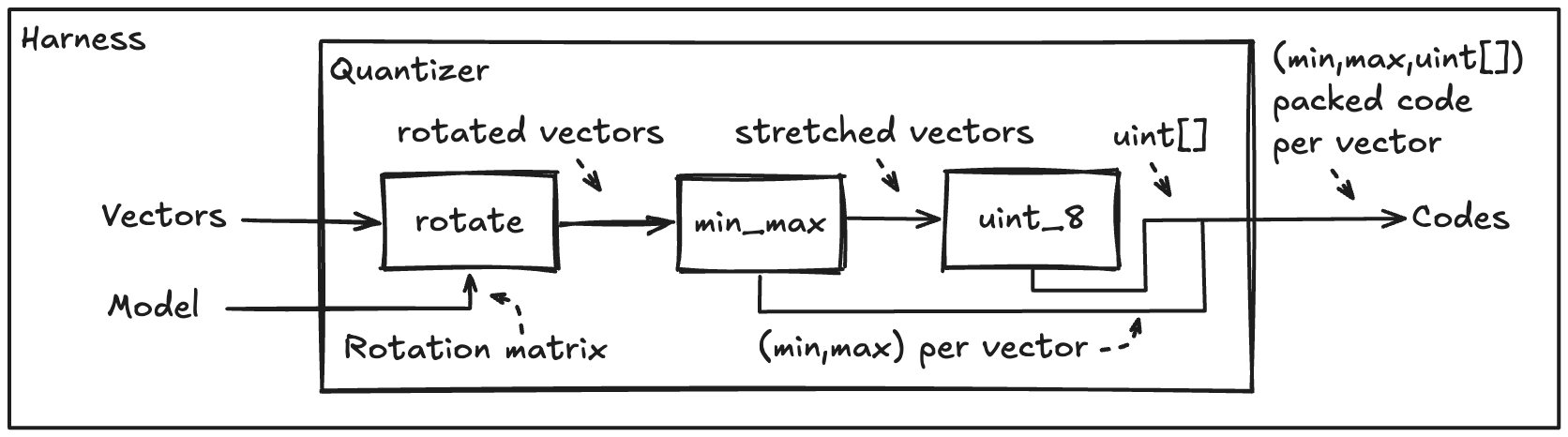}
\caption{An example data flow of \code{encode()} for a quantizer. Specifically, this quantizer is
\code{random\_rotate().adjust(minmax).cast(uint,8)}. The harness chooses and
fetches the data and runs the quantizer with its model \code{encode(model, vectors,...)}.
The quantizer invokes each primitive with its own model (\code{adjust(minmax)} and
\code{cast(uint,8)} don't have models) and applies each primitive's
transformation to generate the next primitive's input vectors. Finally, the
quantizer concatenates the individual primitives' codes (\code{random\_rotate()}
generates no code or side information) into the complete codes for the raw
vectors.}
\label{img:encode}
\end{figure}

Finally, creating a unified framework surfaced many new ideas for improvements.
For example, adaptive encoding can be implemented for measures like KL loss.
Rounders can combine any codebook with any rounding strategy (which motivates new algorithms research). 
Another example is a new primitive, \emph{whitening}, similar to recent papers on covariance-aware rotations \cite{zhou2026oscarofflinespectralcovarianceaware}. 
It equates minimizing scoring error with minimizing reconstruction error and makes the query distribution isotropic. 
Let $U\Sigma^2 U^T= QQ^T$. Map $X' = \Sigma U^TX$ and $Q' = \Sigma^{-1} U^TQ$.
Then, reconstructing the whitened vectors $X'$ gives $\|\tilde{X}'-X'\|_F^2 = \|Q^T(\tilde{X}-X)\|_F^2$. 
A framework like VQ-bench makes it simple to run ablation tests such as prepending a whitening step to any other quantizer.

\subsection{Historical overview}
\label{sec:review}

The literature on vector quantization traces back to Claude Shannon's
rate-distortion theory~\cite{shannon1959coding}, which established that
block source codes (the information-theoretic term for vector quantizers) can
achieve arbitrarily close to optimal performance. VQ remained largely theoretical until the signal processing community developed practical
algorithms. Lloyd and Max independently obtained iterative methods for
optimal \emph{scalar} quantization in the late 1950s and 1960s~\cite{lloyd1982,max1960}. A practical
breakthrough came in 1980 with the LBG algorithm of Linde, Buzo, and
Gray~\cite{linde1980lbg}, which effectively computed $k$-means (Lloyd's
algorithm) on input vectors. This sparked widespread adoption of VQ
in speech and image processing, spurring further
refinement of the underlying techniques~\cite{Gersho82,juang1982multistage,gray1984vq,makhoul1985vector}.
By the early 1990s, most quantization primitive functions - centering, clustering,
hierarchical space-partitioning, lattice coding, dithering - were well
understood and commonly practiced; the classic book by Gersho and
Gray~\cite{Gersho1991VectorQA} reviews the topic circa 1991. Motivated by
signal compression, these works set the target of minimizing reconstruction error.

Minimizing reconstruction error, however, turned out to be too high a bar.
In machine learning, numerical linear algebra, and vector search, it is enough to estimate
$\langle q, x\rangle$ for a set of query vectors $q$.
To see that this is an easier task, consider the unit vector $x \in \R^d$ whose entries are i.i.d.\ $\pm 1/\sqrt{d}$. 
Clearly, representing $\tilde x$ such that $\|x -\tilde x\| \le 1/3$ requires $\Omega(d)$ bits of information.
While the above requires $\Omega(d)$ bits per vector, it gives the strong score approximation guarantee that \emph{for all} unit norm queries $|\langle q, x\rangle - \langle q, \tilde x\rangle| \le 1/3$.
But, we can achieve the same guarantee for an arbitrary set of $m$ queries by encoding $x$ into only $O(\log(m))$ bits.
This was proven in a seminal result by Johnson and Lindenstrauss~\cite{jl1984}. 
It hinges on applying a uniform random rotation to both $x$ and $q$ (which preserves dot products) and then 
computing the approximate dot product using only the first few coordinates of the rotated vectors.\footnote{More commonly known as Random Projection or the JL-lemma.}
Random rotations remain a workhorse of VQ to this day. Over time, random projections were replaced by i.i.d.\
$\pm1$~\cite{achlioptas2003rp} or Gaussian~\cite{dasgupta2003jl} matrices, and
finally made optimal via partial fast Hadamard
transforms~\cite{ailon2009srht,ailon2009dualbch}, requiring only $O(d)$ bits
to store and $O(d\log k)$ operations to apply.

The same probabilistic, reconstruction-free view of compression gave rise to \emph{locality-sensitive hashing} (LSH) for approximate nearest neighbor search;
introduced by Indyk and Motwani~\cite{indyk1998ann} and, independently, by Kushilevitz, Ostrovsky, and Rabani~\cite{KOR2000}. 
This line of work eventually yielded near-optimal constructions~\cite{andoni2006lsh}.
Charikar's SimHash~\cite{charikar2002simhash} is a classic and elegant result in this line of work. 
It is still a top-performing quantizer if tweaked slightly (see \code{hamming} vs.\ \code{sign} in Table~\ref{tab:primitives}).

\subsection{A remark about neural network weight quantization}
\label{sec:qat}
In the context of quantizing neural network weights, techniques are broadly divided into quantization-aware training (QAT) and post-training quantization (PTQ). While both produce quantized weights, QAT is fundamentally a set of \emph{model training} techniques.
It explicitly handles the constraint of gradient steps being quantized, restricting the network to only adjust the codebook or re-assign vectors to different codewords. 
Examples include BinaryConnect~\cite{courbariaux2015binaryconnect},
XNOR-Net~\cite{rastegari2016xnornet}, TWN/TTQ~\cite{li2016twn,zhu2017ttq},
integer-only inference~\cite{jacob2018quant}, PACT~\cite{choi2018pact},
HAWQ~\cite{dong2019hawq}, ProxQuant~\cite{proxquant}, and LSQ
(Learned Step Size)~\cite{esser2020lsq}. These methods are squarely out of the scope of
this paper and of \vqb; QAT methods are omitted from the
quantizer catalog of \S\ref{sec:catalog}.

PTQ, on the other hand, fits \vqb and our framework perfectly. 
In the literature, making VQ ``training aware'' refers to explicitly minimizing score MSE which is $\|Q^T(\tilde{X}-X)\|_F^2$ (see Section~\ref{sec:quality}).
For efficiency reasons, these techniques usually store only a factorization of $QQ^T$ as a surrogate for the Hessian \cite{kim2023squeezellm,pmlr-v119-nagel20a,tseng2024quipsharp,chee2023quip}. Importantly, many PTQ results need $QQ^T$ only for encoding, and not for scoring. 
For model compression, where all the weights are encoded ahead of time, this part of the model can be safely discarded.
Today, \vqb still penalizes such techniques unfairly. It requires that all quantizers are able to encode new vectors (and therefore must keep the Hessian surrogate in their model). This subtlety will be better accounted for in future iterations of \vqb.

\subsection{Best effort disclaimer}
This manuscript and \vqb are provided as a service and resource for the community. 
We have made every effort to be thorough, accurate, and fair. 
We acknowledge, however, that we surely fell short of our ideals on all three.
We almost surely failed to provide the most authoritative citation for some ideas. 
Many quantizers are not yet explained or implemented, and for those covered, we did not describe their proofs, algorithmic nuances, practical insights, and other contributions. 
Finally, \vqb implements all quantizers through its own abstractions, with the goal of being modular.
However, the modularity may be less favorable to some algorithms in terms of running time or memory consumption. 
The framework is implemented in Rust and is high performance enough for benchmarking on real-world size datasets.
It is, however, not the most efficient possible implementation for every algorithm.  
We welcome comments about this manuscript and contributions to \vqb and will do our best to keep both evergreen. 
The benchmarks are made available and kept up to date at \href{https://www.vq-bench.com}{www.vq-bench.com}.
\vqb's codebase, usage instruction, and ways to contribute are at \url{https://github.com/pinecone-io/vq-bench}.

% =====================================================================
\section{Algorithmic primitives}
\label{sec:primitives}

One of the key observations in this paper is that many quantizers
are composed of a rather small set of ideas and algorithmic building blocks, or primitives, that are simply chained together. 

Consider, for example, applying a random rotation to the input vectors.
Downstream processing consumes only rotated vectors, remaining entirely agnostic to the pre-rotation data vectors.
At reconstruction time, after the downstream processing produces a reconstruction of a rotated vector, 
the inverse rotation must be applied to align with the original data vectors.
It is natural to think of random rotation as a functional step in a linear pipeline that transforms data downstream and later back upstream.

We now present the primitives and divide them into four main groups \emph{conditioners}, \emph{rounders}, \emph{splitters}, and \emph{routers}.
The division into groups is semantic and provided for pedagogical reasons only. 
Table~\ref{tab:primitives} lists all \nprimitives\ primitives in four families;
Appendix~\ref{app:primitives} gives more complete definitions.

\subsection{Conditioners} 
Conditioners are invertible (or pseudo-invertible) linear transformations that give
downstream data desired geometric or statistical properties.
For example, centering minimizes the total sum of squared vector norms,
random rotation ensures all coordinate values distribute evenly,
and whitening can make the query distribution isotropic.

Conditioners require an initial fitting step, saving the parameters they need into a model. 
They apply a forward transformation to encode vectors, and an inverse transformation to reconstruct them. 
They apply a complementary transformation to queries to ensure downstream inner products are preserved. 
While most conditioners require only a global model, some must retain per-vector information; 
for example, a normalization step must save the individual vector norms.
Conditioners fall into three conceptual subgroups:
\prg{\code{adjust}} applies a per-coordinate function depending on per-vector or per-coordinate values.  It can emit per-vector parameters as side information. Examples include \code{normalize}, \code{minmax} scaling,  \code{center}, \code{dither}, etc.
\prg{\code{random\_rotate}} applies a data-independent orthogonal map $x\mapsto Rx$; $R$ is stored in the model. Examples include \code{full} rotation, \code{jl} projection, and random \code{hadamard} transform.
\prg{\code{precondition}} applies a data-dependent linear map $x\mapsto Ax$ stored in the model. Examples include learned rotations to minimize quantization error and whitening.

\subsection{Rounders} 
Rounders are defined by two core components: a codebook and a rounding strategy. A codebook (or dictionary) $C \subset \R^d$ is a set of $2^b$ vectors $c \in \R^d$ called codewords. The encoding efficiency, measured in bits-per-dimension, is $\log_2(|C|)/d = b/d$.

Codebooks can be one-dimensional or multi-dimensional, and either fixed or learned from the data. Learned high-dimensional codebooks are more accurate but are also more resource demanding.

Rounders encode vectors by mapping them to a codeword $c$. During reconstruction, they perform a simple lookup, returning $\tilde{x} = c$. While rounders often serve as terminal nodes in a quantization pipeline, they can also be chained together. By default, they emit the residual $x - \tilde{x}$ downstream for further approximation.

The rounding strategy is the algorithm that governs how the rounder assigns a vector $x$ to a codeword $c$. Finding the optimal assignment is often computationally intensive. Consequently, much of the algorithmic innovation in quantization revolve around designing efficient approximation algorithms for rounding. We consider five main types of rounding strategies.

\prg{\code{closest}} returns $\arg\min_{c \in C} \|x-c\|_2$, the default rounding strategy~\cite{linde1980lbg}.

\prg{\code{angular}} returns $\arg\max_{c \in C} \langle x, c \rangle/\|c\|$  \cite{gao2024rabitq,gao2024erabitq}. This is akin to normalizing $C$. During reconstruction, it returns $c$ scaled appropriately.

\prg{\code{random}} (or stochastic rounding) induces a probability distribution $p_x$ over $C$, then samples $c$ according to $p_x$. 
Usually, $p_x$ is designed to minimize variance subject to reconstruction being unbiased, i.e., $\E[\tilde x] = x$. 
Stochastic rounding in one dimension is a simple example of this idea \cite{benbasat2024avq}. 
Similar guarantees can be achieved with \code{closest} rounding preceded by \code{dither}~\cite{feng2026rht}.

\prg{\code{joint}} chooses multiple codewords from multiple codebooks simultaneously. In this context, it solves the following optimization:
\[
{\arg\min}_{(c_1,c_2,\ldots ,c_k) \in (C_1 \times C_2 \times \cdots \times C_k)} \|\textstyle\sum_i Q^T(x- c_i)\|_2
\]
While joint rounding is computationally challenging, recent papers provide encouraging results~\cite{babenko2014aq,martinez2016lsq,egiazarian2024aqlm}.

\prg{\code{adaptive}} is a greedy, sequential implementation of \code{joint} rounding. 
For codebook $i$ it returns $c_i = \arg\min_{c \in C_i}  \|Q^T(x-c) + \delta_i\|_2$, where $\delta_i = \sum_{j<i} Q^T(x- c_j)$ is the accumulated score error~\cite{nagel2020adaround}.
Standard residual encoding is the special case in which $Q = I_d$. 
When the number of sample queries exceeds $d$, one can replace $Q^T$ with an upper triangular $d \times d$ matrix using QR decomposition.

\subsection{Splitters and Routers}
Unlike conditioners and rounders, splitters and routers have multiple downstream branches.
Both are simple constructs and conceptually fit the same interface as conditioners and rounders.
\code{split} partitions each vector and sends one part to each child. During reconstruction, the parts are recombined (e.g. concatenated or added); for scoring, the contributions are summed. 
\code{route} sends each vector to exactly one child and emits the branch identifier. Reconstruction and score defer entirely to that child.
See Appendix~\S\ref{sec:splitters} for details.

\subsection{\vqb's compositional interface}
Arguably, the main contribution of \vqb is the observation that all such
primitives can be captured by a single interface, allowing us to chain and
compose them arbitrarily. All conditioners and rounders implement the following interface:

\begin{lstlisting}
fit(vectors, queries) -> model
encode(model, vectors) -> codes
apply(model, vectors, codes) -> child_vectors
apply_queries(model, queries) -> child_queries
reconstruct(model, codes, child_recons) -> recons
score(model, queries, codes, child_scores) -> scores
\end{lstlisting}

\prg{\code{fit}} produces the internal model from a batch of vectors and
(optionally) queries; the vectors passed to fit are usually a small representative sample of the vectors to be quantized.
\prg{\code{encode}} emits the bit encodings or side information this operator needs to reconstruct or propagate
scores; some primitives emit no codes.
\prg{\code{apply}} transforms the vector
batch for downstream primitives (by default, the residual is passed).
\prg{\code{apply\_queries}} does the same for queries; this is usually either the identity, equal to \code{apply}, or the dual of \code{apply}.
\prg{\code{reconstruct}} is, to
the best extent possible, the inverse of \code{encode} and \code{apply}.
\prg{\code{score}} combines the child's partial scores with its own.

Three practical decisions make \vqb simpler and more efficient. Primitive
functions are \emph{batched}, operating on matrices of vectors, queries, and codes. 
This is semantically identical to a vector-by-vector operation but more efficient. 
Primitives are \emph{neighbor-blind}: the quantizer alone supplies each handle's inputs and consumes its outputs. Finally, \code{encode} and
\code{apply} are deliberately split to simplify the interface.

\begin{table}[htbp]
  \caption{The \nprimitives \vqb primitives and \nprimitivesfunction primitive
   functions. Family-wide default
  contracts are stated in \S\ref{sec:primitives}; full
  \code{fit}/\code{apply}/\code{reconstruct}/\code{score} definitions are in
  Appendix~\ref{app:primitives}.}
  \label{tab:primitives}
  \footnotesize
  \setlength{\tabcolsep}{3pt}
  % Group label spanning the rows of one primitive subgroup (family: subgroup + description).
  \newcommand{\grp}[2]{\multirow[t]{#2}{0.20\textwidth}{\raggedright #1}}
  \begin{threeparttable}
  \begin{tabular}{@{}l >{\ttfamily}l p{0.36\textwidth} l@{}}
    \toprule
    \textbf{Group: Primitive} & \normalfont\textbf{Primitive function}\tnote{a} & \textbf{One-line semantics} & \textbf{Ref.} \\
    \midrule
    \grp{\textbf{Conditioner: \code{adjust}}\\ maps $x\mapsto ax+b$ except for \code{clip}; parameters may be learned or computed per vector and emitted as side information.}{3}
        & \code{adjust(normalize)} & $x\mapsto x/\lVert x\rVert$; $\lVert x\rVert$ emitted as side info & \\
        & \code{adjust(minmax)} & $x\mapsto ax+b$,  $x_i \in [\ell,h]$; $(a,b)$ emitted as side info & \\
        & \code{adjust(absmax)} & $x\mapsto x/\max_j|x_j|$  symmetric, $x_i \in[-1,1]$ & \\
        & \code{adjust(clip)} &   $x\mapsto \max(\min(x,h),\ell)$ for given lower and upper bounds; lossy. $\ell \le h$  & \cite{badri2024hqq} \\
        & \code{adjust(normalize\_coords)} & $x \mapsto s^{-1} \circ x$ where $s$ are coordinate norms  & \cite{lin2023awq,xiao2023smoothquant} \\
        & \code{adjust(dither)} & $x\mapsto x+\epsilon$, $\epsilon\sim\psi$ & \cite{roberts1962,KiracDither,feng2026rht} \\
        & \code{adjust(center)} & $x\mapsto x-\mu$; \code{score} adds the per-query constant $\langle q,\mu\rangle$ & \\
    \midrule
    \grp{\textbf{Conditioner: \code{random\_rotate}}\\ $X\mapsto RX$, $Q\mapsto RQ$, $\tilde{X}\mapsto R^T\tilde{X}$; $R$ is data-independent}{4}
        & \code{random\_rotate(full)} & $x\mapsto Rx$; $R$ is uniform rotation;  $O(d^2)$ space and time & \\
        & \code{random\_rotate(jl)} & $x\mapsto Rx$; $R$ is uniform $k$-dimensional subspace; $O(dk)$ space and time & \cite{jl1984,dasgupta2003jl,achlioptas2003rp,matousek2008jl} \\
        & \code{random\_rotate(hadamard)} & $x\mapsto HDx$, $H$ Hadamard, $D$ random $\pm1$ diagonal. & \cite{ailon2009srht} \\
        & \code{random\_rotate(partial\_hadamard)} & $x\mapsto PHDx$; Hadamard evaluated at $k$ coordinates; $O(d\log k)$ time. & \cite{ailon2009dualbch} \\
    \midrule
    \grp{\textbf{Conditioner: \code{precondition}}\\ Optimizes to find $A$. Store $A$ in the model. Then $ X \mapsto AX$, $Q \mapsto (A^{\dagger})^{T}Q$, $\tilde{X} \mapsto \tilde A^{\dagger}X$}{5}
        % & \code{precondition(pca)} & compute svd $X=U\Sigma V^T$; $ X \mapsto U^T_k X$; lossy when $k<d$ & \\
        & \code{precondition(whitening)} & compute svd $U\Sigma^2 U^T \approx \E[qq^T]$; $X \mapsto \Sigma U^T X $, $Q \mapsto \Sigma^{-1}U^TQ$ & This paper\tnote{b} \\
        & \code{precondition(optimize\_signs)} & $ X \mapsto RX$ minimizing reconstruction cost for \code{sign}  & \cite{gong2013itq,ge2013opq} \\
        & \code{precondition(optimize\_pq)} & $ X \mapsto RX$ minimizing reconstruction cost for \code{pq} & \cite{gong2013itq,ge2013opq} \\
        & \code{precondition(optimize\_int)} & $ X \mapsto RX$ minimizing reconstruction cost for \code{int} &  \cite{liu2025spinquant} \\
    \midrule
    \grp{\textbf{Rounder: \code{cast}}\\ map to a data agnostic set of $2^b$ fixed codewords; no model; reconstruct up-casts}{8}
        & \code{cast(fp16)}, \code{cast(fp8)}, \code{cast(fp4)} & IEEE half-precision round; \code{E4M3} or \code{E5M2}; \code{E2M1} or \code{E1M2} & \cite{micikevicius2022fp8} \\
        & \code{cast(uint,b)} , \code{cast(int,b)} & $b$-bit integer lattice; implicitly preceded by \code{minmax} to the range.  & \cite{gray1998quant} \\
        % & \code{cast(sign)} & one bit; $\pm1$ values; symmetric scoring by hamming, asymmetric by default. & \cite{charikar2002simhash}\cite{gordo2014asymmetric}\cite{gao2024rabitq} \\
        & \code{cast(sign)} & $\{-1,1\}$: viewed as real values \code{score}$(q,s)=\sum_i q_i s_i$ (asymmetric) & \cite{gordo2014asymmetric}\cite{gao2024rabitq} \\
        & \code{cast(hamming)} & $\{0,1\}$: $1$ iff $x \ge 0$; \code{score}$=d-2\,\mathrm{popcount}(\mathrm{apply}(q)\oplus b)$ & \cite{charikar2002simhash} \\
        & \code{cast(ternary)} & $x_i\mapsto 0$ if $|x_i|<1$, else $\mathrm{sign}(x_i)$ & \cite{li2016twn}\cite{zhu2017ttq} \\
        & \code{cast(nf4)}, \code{cast(beta)}& fixed 1d codebooks for normal and beta distributed variables & \cite{dettmers2023qlora} \\
        & \code{cast(MX)} & microscaling formats (applied to blocks) & \cite{rouhani2023mx} \\
        & \code{cast(E8)} & fixed optimal data-agnostic codebook in 8-dimensions & \cite{tseng2024quipsharp} \\
    \midrule
    \grp{\textbf{Rounder: \code{kmeans}}\\ explicitly train and save a table of $2^b$ codewords, stored as the model; reconstruct/score by lookup table.}{5}
        & \code{kmeans(lloyd)} & Lloyd's algorithm; encode by nearest centroid & \cite{linde1980lbg,lloyd1982,Gersho1991VectorQA} \\
        & \code{kmeans(random\_round)} & a 1-d specialized optimization to reduce stochastic rounding  variance & \cite{benbasat2024avq} \\
        & \code{kmeans(angular)} & minimize clustering cost in cosine-distance & \\
        & \code{kmeans(anisotropic)} & \code{fit} minimizes a \emph{score} loss; otherwise shaped like \code{kmeans} (ScaNN) & \cite{guo2020scann} \\
    \midrule
    \grp{\textbf{Splitter: \code{split}}\\ A segment per child; \code{reconstruct} concatenates/adds, \code{score} sums}{3}
        & \code{split(segment)} & fixed number of coordinates per segment & \cite{jegou2011pq} \\
        % & \code{split(variable\_segments)} & variable coordinates per segment, balancing $\ell_2^2$ energy per segment & \cite{Gersho1991VectorQA} \\
        & \code{split(outliers)} & \code{fit} finds outlier coordinates; split to two segments, outliers and the rest & \cite{dettmers2022llmint8} \\
        & \code{split(additive)} & full-width sub-codebooks; joint fit & \cite{babenko2014aq,martinez2016lsq,egiazarian2024aqlm} \\
        & \code{split(sparse\_dense)} & a sparse stream (bounded $\|x\|_0$) plus a dense stream (bounded $\|x\|_\infty$) & \cite{kim2023squeezellm,kang2024gear} \\
    \midrule
    \grp{\textbf{Router: \code{route}}\\ send each vector to one child}{2}
        & \code{route(outliers)} & branch on $\lVert x\rVert$ (one or several norm bins) & \\
        & \code{route(kmeans)} & $k$-means as coarse clustering; route to nearest centroid, emit branch id & \cite{jegou2011pq} \\
    \bottomrule
  \end{tabular}
  \begin{tablenotes}[flushleft]\footnotesize
  \item[a] The primitive names used in this table do not necessarily align with those used in the \vqb implementation.
  \item[b] Whitening as a whole is obviously not a new idea; the authors, however, could not find instances of it being
  applied to improve quantization.
  \end{tablenotes}
  \end{threeparttable}
\end{table}

% =====================================================================
\section{Quantizers: primitive composition and execution}
\label{sec:quantizers}

Using the interface above, quantizers compose and execute pipelines of primitives.
See Appendix~\ref{sec:chaining} for more details about chaining operations and notation.
Intuitively, the following \code{PQ = split(segment,width=4)\allowbreak.kmeans(k=256)}
should read as: split every vector into contiguous segments of 4 coordinates, then feed each
sub-vector to an independent \code{kmeans} rounder to create an 8-bit vector lookup table.

Quantizers use the primitive interface above to implement complete pipelines with the following interface. 
See Appendix~\ref{app:pseudocode} for exact breakdown.

\begin{lstlisting}
fit(vectors, queries) -> model
encode(model, vectors) -> codes
reconstruct(model, codes) -> recons 
score(model, queries, codes) -> scores
\end{lstlisting}
%reconstruct(model, codes, child_recons) -> recons
%score(model, queries, codes, child_scores) -> scores

\subsection{Worked example: centered SimHash}
\label{sec:worked}
A full trace of \code{center().random\_rotate().hamming()}.

\vspace{0.2cm}
\noindent \prg{\code{fit(vectors)}} the pipeline runs \code{center.fit} $\to$
  applies to get $x-\mu$ $\to$ \code{random\_hadamard.fit} $\to$ applies to
  get $R_H(x-\mu)$ $\to$ \code{hamming.fit} (nothing to learn).

\prg{\code{encode(vectors)}} \code{center} $\to$ ($\varnothing$, $x-\mu$);
  \code{random\_hadamard} $\to$ ($\varnothing$, $R_H(x-\mu)$);
  \code{hamming} $\to$ $b=\operatorname{sign}(R_H(x-\mu))$, residual dropped.
  Return (to the harness) the bit encoding $b$.

\prg{\code{score(queries, codes)}} the forward map gives $q$ for
  \code{center} and $R_H q$ for \code{random\_hadamard};
  \code{hamming.score} returns
  $\hat s = d - 2\operatorname{pop count}(\operatorname{sign}(R_H q)\oplus b)$;
  folding back, \code{center} adds $\langle q,\mu\rangle$.

\prg{\code{reconstruct(codes)}}: \code{hamming} $\to\pm1$;
  \code{random\_hadamard} $\to R_H^T b$; \code{center} $\to R_H^T b+\mu$.

\subsection{Quantizers catalog}
\label{sec:catalog}

Table~\ref{tab:catalog} surveys \nquantizers\ published quantizers, each as an explicit pipeline over the primitives of \S\ref{sec:primitives}. Writing all methods in one vocabulary helps discover commonalities between quantizers; most new methods differ from older ones in preconditioning and rounding strategies, not in their structure. For instance, OPQ adds a jointly trained rotation to PQ; LSQ (local search) swaps
AQ's rounding strategy to beam search; HQQ fits \code{cast(uint,b)}'s scale by
half-quadratic optimization. These are exactly the kind of
distinctions a common framework is designed to surface and measure.

The catalog is split into three tiers: methods implemented in VQ-bench, methods supported by the framework but not yet implemented, and methods not supported.

\begin{table}[htbp]
  \caption{\nquantizers\ published quantizers as explicit pipelines over the primitives
  of Table~\ref{tab:primitives}, in three sections: \ncatalogimplemented\ implemented in \vqb; \ncatalogsupported\ supported by the framework but not yet
  implemented; and \ncatalogunsupported\ not supported.}
  \label{tab:catalog}
  \footnotesize
  \setlength{\tabcolsep}{3pt}%
  \let\code\codepipe
  \begin{tabular}{@{}l c >{\ttfamily\raggedright\arraybackslash}p{0.37\textwidth} >{\raggedright\arraybackslash}p{0.27\textwidth} l@{}}
    \toprule
    \normalfont\textbf{Method} & \textbf{Year} & \normalfont\textbf{Pipeline} & \textbf{Note} & \textbf{Ref.} \\
    \midrule
    \multicolumn{5}{@{}l}{\normalfont\emph{Implemented in \vqb}} \\
    SimHash & 2002 & \code{random\_rotate(jl).cast(hamming)} & Hamming distance estimates angle & \cite{charikar2002simhash} \\
    PQ & 2011 & \code{split(segment).kmeans(lloyd)} & per-segment scores are summed & \cite{jegou2011pq} \\
    ITQ & 2013 & \code{precondition(optimize\_signs).cast(hamming)} & rotation minimizes sign-quantization error & \cite{gong2013itq} \\
    OPQ & 2013 & \code{precondition(optimize\_pq).split(segment).kmeans(lloyd)} & rotation minimizes PQ-reconstruction error & \cite{ge2013opq} \\
    ITQ, asymmetric & 2014 & \code{precondition(optimize\_signs).cast(sign)} & continuous query scored against signs & \cite{gordo2014asymmetric} \\
    EDEN / EDEN-unbiased & 2022 & \code{adjust(normalize).random\_rotate(hadamard).cast(beta,b)} & differ only in per-vector scaling$^\dagger$ & \cite{vargaftik2022eden} \\
    QJL & 2024 & \code{random\_rotate(jl).cast(sign)} & unbiased inner product estimate & \cite{zandieh2024qjl} \\
    RaBitQ & 2024 & \code{adjust(center).adjust(normalize).random\_rotate(hadamard).cast(int,1,angular)} & per-vector scale guarantees unbiasedness & \cite{gao2024rabitq} \\
    E-RaBitQ & 2024 & \code{adjust(center).adjust(normalize).random\_rotate(hadamard).cast(int,b,angular)} & $b$-bit extension of RaBitQ & \cite{gao2024erabitq} \\
    TurboQuant & 2025 & \code{adjust(normalize).random\_rotate(hadamard).cast(beta,b-1).random\_rotate(hadamard).cast(sign)} & reallocates one bit for unbiased scoring & \cite{zandieh2025turboquant,zandieh2024qjl} \\
    \midrule
    \multicolumn{5}{@{}l}{\normalfont\emph{Supported, not yet implemented}} \\
    RVQ & 2010 & \code{kmeans(lloyd). ... .kmeans(lloyd)} repeated $\ell$ times & each level quantizes the prior residual & \cite{chen2010rvq,juang1982multistage} \\
    AdaRound & 2020 & \code{cast(int, b, adaptive)} & learns up-down rounding decision via SGD & \cite{nagel2020adaround} \\
    GPTQ & 2023 & \code{cast(int,b,adaptive)} & adaptive rounding to integer lattice & \cite{frantar2023optq} \\
    NF4 / QLoRA & 2023 & \code{split(segment).adjust(absmax).cast(nf4)} & QLoRA double-quantizes the block scales & \cite{dettmers2023qlora} \\
    QuIP & 2023 & \code{random\_rotate(hadamard).cast(int,b,adaptive)} & omitting rotation recovers GPTQ & \cite{chee2023quip} \\
    %AWQ & 2023 & \normalfont not implemented yet & calibration grid-searched per-channel scale & \cite{lin2023awq} \\
    QuaRot & 2024 & \code{random\_rotate(hadamard).cast(int,b)} & applies rotation to remove outliers & \cite{ashkboos2024quarot} \\
    QuIP\# & 2024 & \code{random\_rotate(hadamard).cast(E8,adaptive)} & QuIP with $2$-bit $E_8$ lattice$^\dagger$ & \cite{tseng2024quipsharp} \\
    KIVI & 2024 & \code{split(segment).cast(uint,b)} & per-segment casting$^\dagger$ & \cite{liu2024kivi} \\
    GEAR & 2024 & \code{split(outliers)[cast(uint,b), precondition(pca,k).cast(fp16)]} & PCA of the final residual & \cite{kang2024gear} \\
    HQQ & 2024 & \code{adjust(hqq).cast(uint,b)} & half-quadratic-optimized zero-point$^\dagger$ & \cite{badri2024hqq} \\
    %KVQuant & 2024 & \normalfont not implemented yet & sensitivity-weighted shared codebook$^\dagger$ & \cite{hooper2024kvquant} \\
    SpinQuant & 2024 & \code{precondition(optimize\_int).cast(int,4)} & learns rotation via Riemannian SGD$^\dagger$ & \cite{liu2025spinquant} \\
    Dithered RHT & 2026 & \code{random\_rotate(hadamard).adjust(dither).cast(int,b)} & unbiased; dithered fast Hadamard$^\dagger$ & \cite{feng2026rht} \\
    \midrule
    \multicolumn{5}{@{}l}{\normalfont\emph{Not supported}} \\
    AQ & 2014 & \code{split(additive,m).kmeans(lloyd,joint)} & joint encode across additive codebooks breaks the stage interface & \cite{babenko2014aq} \\
    LSQ (local search) & 2016 & \code{split(additive,m).kmeans(lloyd,joint)} & beam search over AQ's joint codes & \cite{martinez2016lsq} \\
    AQLM & 2024 & \code{split(additive,m).kmeans(lloyd,joint)} & joint codebook fit and encode & \cite{egiazarian2024aqlm} \\
    \bottomrule
  \end{tabular}

  \smallskip
  {\footnotesize $^\dagger$\,Expanded in Appendix~\ref{sec:quantizer-detail}.}
\end{table}

% =====================================================================
\section{The \vqb harness}
\label{sec:harness}

The \vqb harness is a program that sits above quantizers and is responsible
for running experiments comprehensively and fairly. The harness hands the
quantizer vectors and queries to \code{fit}, then vectors to \code{encode}; afterwards it
asks it to \code{reconstruct} certain vectors and to \code{score} queries
against candidate vectors. Crucially, a quantizer is never responsible for
measuring its own quality: the harness evaluates the size of the encoding and
the quality of the reconstructions and approximated scores. Each experiment
consists of three components: the datasets (and their evaluation parameters),
the quantizers (and their parameter settings), and the metrics to be
evaluated.

\prg{Datasets.}
Datasets are HDF5 files with four arrays.
\emph{base}: database vectors,
\emph{eval}: evaluation queries,
\emph{eval\_candidates}: top-$L$ neighbors of each evaluation query,
and an optional \emph{calib}: held-out calibration queries available to \code{fit}.

\vqb maintains a registry of dot-product
datasets from ANN-Benchmarks~\cite{aumuller2017annbench} and
VIBE~\cite{jaasaari2025} and reformats them accordingly. All vector coordinates are stored in \code{float32} precision.

\prg{Experiment specification.}
An experiment is a single JSON file specifying datasets, quantizers, and
metrics. For example, the following evaluates three quantizers on two
datasets, measuring recall@$\{1,10,100\}$ and MSE of
scores and reconstructions:

\begin{lstlisting}
{
  "datasets": ["arxiv-nomic-768-normalized",
               "yahoo-minilm-384-normalized"],
  "seed": 1,
  "n_fit": 50000,
  "n_reconstruct": 1000,
  "n_eval": 1000,
  "k": [1, 10, 100],
  "methods": [
    { "name": "pq", "centroids": 256,
                    "section_dim": [8, 4, 2] },
    { "name": "minmax", "b": [1, 2, 4, 6] },
    { "name": "rabitq" }
  ],
  "metrics": ["recall", "mse_score", "mse_recon"]
}
\end{lstlisting}
Here, \code{n\_fit} sets how many randomly sampled database vectors are handed to
\code{fit}, \code{n\_reconstruct} how many the harness asks each quantizer to
\code{reconstruct}, and \code{n\_eval} how many queries it passes to \code{score}.
The random seed used to draw all three samples is given by \code{seed}.

\prg{Execution.}
For each dataset, the harness loads the database into \code{vectors}, the
\code{n\_fit} sampled vectors to fit on into \code{fit\_vectors}, the
indices to reconstruct into \code{recon\_indices}, the queries to score into
\code{queries}, and their top-$L$ neighbors into \code{candidates}. For each quantizer
and parameter set it executes:
\begin{lstlisting}
model  = quantizer.fit(fit_vectors, calib)
codes  = quantizer.encode(model, vectors)
recons = quantizer.reconstruct(model, codes[recon_indices])
scores = quantizer.score(model, queries, codes[candidates])
\end{lstlisting}
Note that \code{fit} receives the \emph{calibration} queries, never the evaluation
queries. The harness also evaluates \code{true\_recons} and \code{true\_scores} for the
selected indices and queries. It then computes compression metrics from
(\code{model}, \code{codes}), and quality metrics from
(\code{true\_recons}, \code{recons}) and (\code{true\_scores}, \code{scores}).

\prg{Storage measurement: bits per dimension.}
To report storage cost, the harness measures the sizes of the returned
\code{model} (learned scalings, rotations, codebooks, \ldots) and \code{codes}
(per-vector encodings). The sum of the sizes in bits divided by $nd$ gives the
effective bits per dimension. The unquantized baseline is 32 bits per
dimension, so $b$ bits per dimension is a compression ratio of $32/b$.

\subsection{Quality measurements}
\label{sec:quality}

\vqb measures a variety of quality metrics because no single metric applies to
all downstream use cases. We use the following notation:
\begin{itemize}
   \item Write $X,\tilde{X}\in\R^{n_r\times d}$ for
  \code{true\_recons} and \code{recons}, where $n_r=\code{n\_reconstruct}$
  \item Write $S,\tilde{S}\in\R^{n_q\times L}$ for \code{true\_scores} and \code{scores}, where $n_q=\code{n\_eval}$ and $L = $ number of candidates per query
  \item $s_{ij}$ and $\tilde s_{ij}$ are the true and approximate scores of candidate $j$ for query $i$.
  \item $T_k(i)$ and $\tilde T_k(i)$ are the candidates with the $k$ highest true and approximate scores for query $i$
\end{itemize}

Model compression metrics include reconstruction loss $\|X - \tilde X\|_F^2$. 
This is still the relevant metric for post-training model weights and KV-cache value compression. 
Recent PTQ quantizers focus explicitly on Score MSE which is $\|Q^T(X - \tilde X)\|_F^2$. 

Vector database metrics capture how well quantization preserves high scores
and rankings, analogous to metrics in traditional search systems. Recall@$k$
is the de-facto ANN metric~\cite{aumuller2017annbench}. 
But, it is too sensitive to arbitrary tie breaking and exhibits other numerical instabilities. 
SOS$_k$ is its numerically stable counterpart. Score bias matters when scores
are aggregated across many vectors, e.g.\ in re-ranking.

KV-cache metrics account for computing the softmax distributions. 
Score MSE weights every error equally but when scores pass through a softmax errors on larger scores are amplified. 
For each query $i$ the softmax maps the $L$ candidate scores to a distribution
$p_{ij} = e^{s_{ij}/\tau} / \sum_{j'} e^{s_{ij'}/\tau}$ with temperature
$\tau$, and the approximate scores give $\tilde p_i$. We measure KL divergence
and total variation between $p_i$ and $\tilde p_i$; TV directly bounds the
\emph{attention error} - if $\lVert p_i-\tilde p_i\rVert_1\le\epsilon$ and
the value vectors have norm $\le 1$, the approximated attention output has
error norm $\le \epsilon$.

\begin{table}[t]
  \caption{Metrics computed by the harness: nine quality metrics from
  $(X,\tilde X)$ and $(S,\tilde S)$, plus the resource measurements it records
  for every run.}
  \label{tab:metrics}
  \centering
  \footnotesize
  \begin{tabular}{@{}l l@{}}
    \toprule
    \textbf{Metric} & \textbf{Definition} \\
    \midrule
    \multicolumn{2}{@{}l}{\emph{Model compression, Post training quantization}} \\
    Reconstruction MSE & $\tfrac{1}{n_r}\sum_i \lVert x_i-\tilde x_i\rVert^2$ \\
    Score MSE & $\tfrac{1}{n_q L}\sum_{i,j}(s_{ij}-\tilde s_{ij})^2$ \\
    Reconstruction bias & $\lVert \tfrac{1}{n_r}\sum_i (x_i-\tilde x_i)\rVert^2$ \\
    Score bias & $\tfrac{1}{n_q L}\sum_{i,j}(s_{ij}-\tilde s_{ij})$ \\
    \midrule
    \multicolumn{2}{@{}l}{\emph{Vector database, semantic search}} \\
    Recall@$k$ & $\tfrac{1}{n_q}\sum_i |T_k(i)\cap\tilde T_k(i)|/k$ \\
    SOS@$k$ (sum of top scores) & $\bigl(\sum_i\sum_{j\in\tilde T_k(i)} s_{ij}\bigr)/\bigl(\sum_i\sum_{j\in T_k(i)} s_{ij}\bigr)$ \\
    \midrule
    \multicolumn{2}{@{}l}{\emph{KV cache (softmax at temperature $\tau$)}} \\
    KL divergence & $\tfrac{1}{n_q}\sum_i D_{\mathrm{KL}}(p_i\,\Vert\,\tilde p_i)$ \\
    TV distance & $\tfrac{1}{2n_q}\sum_i\sum_j |p_{ij}-\tilde p_{ij}|$ \\
    exp-SOS@$k$ & $\bigl(\sum_i\sum_{j\in\tilde T_k(i)} e^{s_{ij}/\tau}\bigr)/\bigl(\sum_i\sum_{j\in T_k(i)} e^{s_{ij}/\tau}\bigr)$ \\
    \midrule
    \multicolumn{2}{@{}l}{\emph{Resource consumption}} \\
    Bits per dimension & total, and split into model and codes \\
    Encode time & wall clock for \code{encode} \\
    Encode memory & peak heap\\
    Score time & per query: mean, p50, p90, p99 \\
    Reconstruct time & per vector \\
    \bottomrule
  \end{tabular}
\end{table}

% =====================================================================
\section{Experimental results}
\label{sec:results}

We evaluated \nimplemented\ quantizers (Table~\ref{tab:quantizers}) on 5 high-dimensional embedding datasets
(Table~\ref{tab:datasets}). In every experiment the harness fits on $50{,}000$
sampled database vectors, then draws $1{,}000$ query samples and $1{,}000$
reconstruction samples and scores $L=1{,}000$ candidates per query.
We measure all the metrics of
\S\ref{sec:quality}: bits per dimension, reconstruction MSE and bias, score
MSE and bias, recall@$k$, SOS@$k$ and exp-SOS@$k$ for $k\in\{1,10,100\}$, and KL divergence
and TV distance for $\tau\in\{0.01,0.05,0.1\}$. Two of the quantizers are baselines:
MinMax applies a per-vector affine transformation followed by integer casting
(\code{adjust(minmax).cast(uint,b)}), and \code{scalar} is the per-coordinate
version of the MinMax. All runs use a single seed
($\code{seed}=1$). Time and memory statistics were measured on an Apple M2 Pro with 16GB RAM, with \code{encode} run on 6 threads.

In this manuscript we include results for all five datasets
and four metrics: reconstruction MSE, score MSE, recall@$10$, and TV
distance at $\tau=0.05$. Figure~\ref{fig:results-arxiv} shows the four metrics
on ArXiv; the grids for the other datasets are presented in
Appendix~\ref{app:plots} (Figures~\ref{fig:results-ccnews}-\ref{fig:results-laion}).

To explore the results themselves, it is better to interact directly with the companion website \url{www.vq-bench.com}.
Some high level insights include the following. Basic approaches like MinMax and SimHash are very fast and memory efficient. But, they fall significantly behind on quality metrics. Other than extreme latency/quality tradeoffs, they are probably not recommended. Optimized approaches like ITQ and OPQ  perform well in the low bit budget regime (1 bit/dim) but pay a penalty in implementation complexity, training time, and compute and memory consumption. Standard PQ and TurboQuant (and their variants) perform well but do not match the top results. In our experiments, the most consistently performing algorithms are EDEN and E-RaBitQ. Not only do they often get the best (or almost the best) compression ratios, they are also quite memory and compute efficient. EDEN has the additional benefit of fast encoding time. Both algorithms are good candidates for most quantization applications.

\begin{table}[t]
\caption{\nimplemented\ evaluated quantizers and their parameter settings.}
\label{tab:quantizers}
\centering
\footnotesize
\setlength{\tabcolsep}{2.5pt}
\begin{tabular}{@{}llc@{}}
\toprule
\textbf{Quantizer} & \textbf{Parameters} & \textbf{Ref.} \\
\midrule
RaBitQ               & None ($1$ bit)                                & \cite{gao2024rabitq} \\
E-RaBitQ             & $b \in \{2,4,6\}$                             & \cite{gao2024erabitq} \\
SimHash              & $b\in\{0.5,1,2,4,6\}$                                          & \cite{charikar2002simhash} \\
ITQ                  & None                                          & \cite{gong2013itq} \\
ITQ-asym             & None                                          & \cite{gordo2014asymmetric} \\
PQ                   & $\code{centroids}=256$, $\code{section\_dim}\in \{8,4,2\}$ & \cite{jegou2011pq} \\
OPQ                  & $\code{centroids}=256$, $\code{section\_dim}\in \{8,4,2\}$ & \cite{ge2013opq} \\
QJL                  & $b\in\{0.5,1,2,4,6\}$                                          & \cite{zandieh2024qjl} \\
EDEN-MSE             & $b\in\{1,2,4,6\}$                             & \cite{vargaftik2022eden} \\
EDEN-prod            & $b \in \{1,2,4,6\}$                           & \cite{vargaftik2022eden} \\
TurboQuant-MSE       & $b \in \{1,2,4,6\}$                           & \cite{zandieh2025turboquant} \\
TurboQuant-prod      & $b \in \{2,4,6\}$                             & \cite{zandieh2025turboquant} \\
Scalar               & $b\in\{1,2,4,6\}$                             & \\
MinMax               & $b\in\{1,2,4,6\}$                             & \\
\bottomrule
\end{tabular}
\end{table}

\begin{table}[t]
\caption{Datasets included in the sweep, all from VIBE~\cite{jaasaari2025};
dot-product similarity throughout. \textbf{Calib.} is the number of held-out
calibration queries used during fitting.}
\label{tab:datasets}
\centering
\footnotesize
\begin{tabular}{@{}llccc@{}}
\toprule
\textbf{Dataset} & \normalfont\code{name} & \textbf{Vectors} & \textbf{Dim.} & \textbf{Calib.} \\
\midrule
ArXiv  & \code{arxiv-nomic-768-normalized}  & $1{,}344{,}643$ & $768$ & $0$ \\
CCNews & \code{ccnews-nomic-768-normalized} & $495{,}328$     & $768$ & $0$ \\
Yahoo  & \code{yahoo-minilm-384-normalized} & $677{,}305$     & $384$ & $0$ \\
COCO   & \code{coco-nomic-768-normalized}   & $282{,}360$     & $768$ & $114{,}054$ \\
LAION  & \code{laion-clip-512-normalized}   & $1{,}000{,}448$ & $512$ & $999{,}448$ \\
\bottomrule
\end{tabular}
\end{table}

\resultgrid{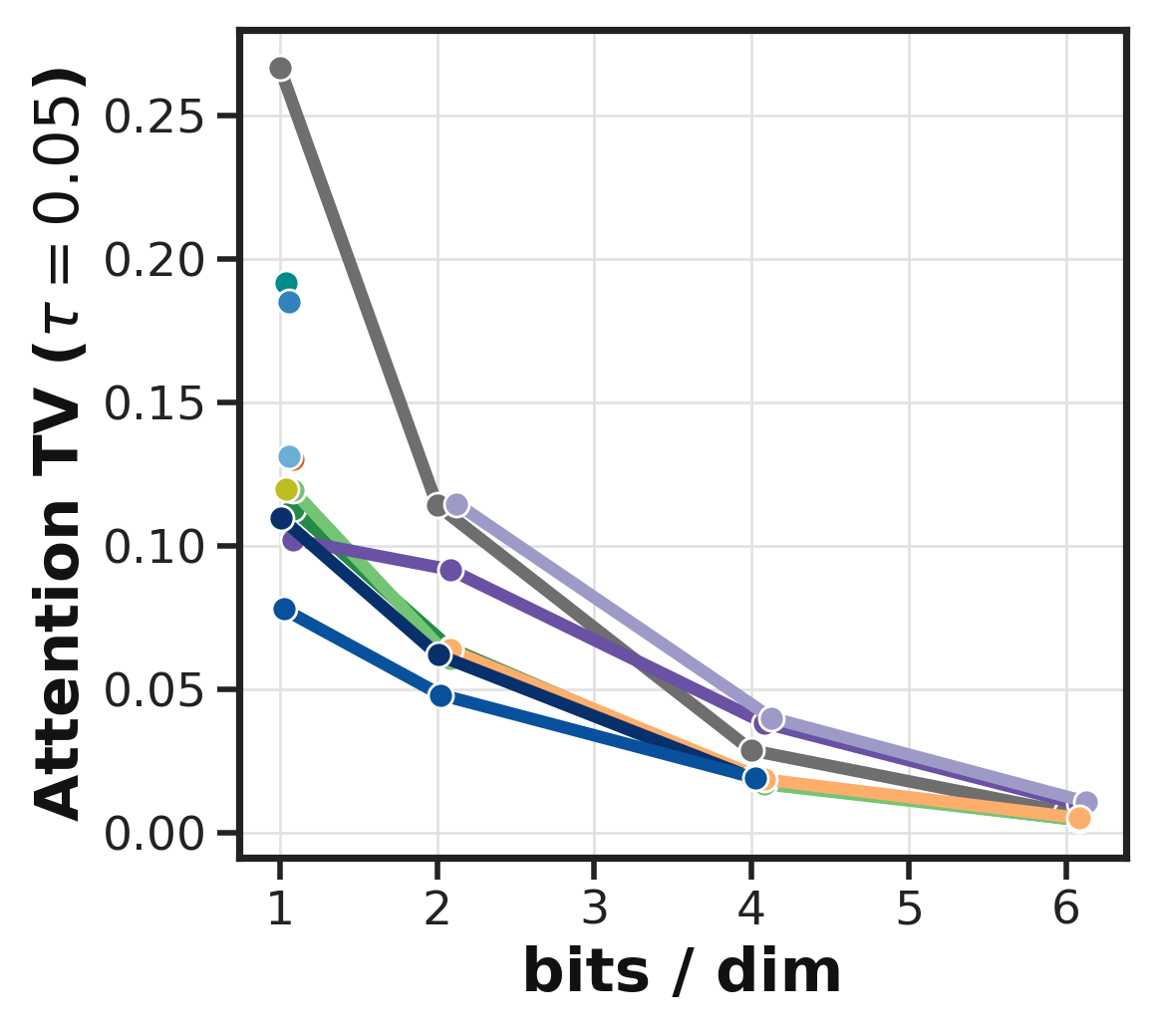}{The four headline metrics versus bits per dimension on
  ArXiv. For detail, MinMax is omitted from Attention
  TV distance.}{fig:results-arxiv}

\FloatBarrier

% =====================================================================
\section{Discussion and future work}
\label{sec:discussion}

We present \vqb, an open-source quantization benchmark that provides a common framework for comparing quantization algorithms, both in terms of their compositional structure and their empirical performance. However, our benchmarking approach is not without its limitations. We do not heavily optimize running times or memory usage: we favor composability and extensibility, and where extreme performance is critical, we expect engineers to reimplement their chosen quantizers without the overhead of abstraction layers. We do not expect to have implemented all methods perfectly, and we solicit feedback and contributions from the authors of those results. We made an attempt to attribute credit correctly but expect to have done so imperfectly, and we welcome corrections. Further limitations include the unfair accounting of PTQ models (\S\ref{sec:qat}) and the fact that evaluation is done on a single machine. Future work includes adding more primitives, datasets, and optimized kernels; searched or learned pipelines over the primitive space; and more extensive ablation experiments.

% =====================================================================
\bibliographystyle{plain}
\bibliography{references}

% =====================================================================
\appendix
% =====================================================================

\section{Primitive catalog (full)}
\label{app:primitives}

This appendix expands Table~\ref{tab:primitives} with the full per-primitive
definitions, organized by family.

Conditioners are invertible (or pseudo-invertible) linear maps that give
downstream vectors a desired geometric or statistical property. \code{fit}
learns and stores the transform; \code{apply} maps the database vectors,
\code{apply\_queries} applies the complementary (dual) map so that dot
products are preserved downstream, and \code{reconstruct} applies the inverse.
Unless otherwise specified, \code{encode} emits no codes and \code{score} is a
pass-through. The three families below differ in how the map is obtained:
fixed per-vector affine (\code{adjust}), data-independent random orthogonal
(\code{random\_rotate}), or learned by optimization (\code{precondition}).

\subsection{Conditioner: \code{adjust}}
The \code{adjust} family applies a per-vector affine map $x\mapsto ax+b$ and
may emit the chosen scalars (or the norm) as side information.

\noindent \prg{\code{adjust(normalize)}} \code{encode} emits $\lVert x\rVert$ as
side information (potentially in low precision); $apply(x)=x/\lVert x\rVert$
maps every vector onto the unit sphere; \code{apply\_queries} is the identity;
\code{score} and \code{reconstruct} multiply back by the stored scalar
$\lVert x\rVert$. Separating magnitude from direction lets a downstream rounder
spend all of its bits on the unit-norm direction; it is the first stage of the
EDEN/TurboQuant/RaBitQ family.

\prg{\code{adjust(minmax)}} bounds $\ell,h\in\R$; \code{fit} computes
$(a,b)$ mapping the observed range affinely into $[\ell,h]$ and emits $(a,b)$
as side information.\footnote{$a = (h-\ell)/(\max_x-\min_x)$ and
$b = (\ell\cdot\max_x - h\cdot\min_x)/(\max_x-\min_x)$. Clearly $(a,b)$
themselves can be saved in low precision if needed.} Then
$apply(x) = ax+b$, $apply\_queries(q) = q$,
$reconstruct(\tilde x) = (\tilde x - b)/a$, and
$score(q,(a,b),\tilde s) = (\tilde s - b\langle q,1\rangle)/a$.
\code{adjust(minmax)} is straightforward and very common in model compression;
learned or calibrated clip thresholds that shrink $[\ell,h]$ to suppress
outliers are a common refinement (e.g.\ PACT)~\cite{choi2018pact}.

\prg{\code{adjust(absmax)}} the symmetric, zero-point-free special case
of \code{adjust(minmax)} - scale by $1/\max_j|x_j|$ so
$x \mapsto x/\max_j|x_j| \in [-1,1]$ with $0\mapsto 0$ (one stored scale, no
offset $b$). Common for per-block weight normalization (e.g.\ NF4).

\prg{\code{adjust(clip)}} 
May be used to remove outlier values.
It gets a pair of values $(\ell,h)$. 
It rounds all values lower than $\ell$ to $\ell$ and all values higher than $h$ to $h$.
HQQ computes $(\ell,h)$ per vector to reduce downstream integer rounding error \cite{badri2024hqq}.
For large outlier values \code{split(sparse\_dense)} is preferable \cite{hooper2024kvquant}

\prg{\code{adjust(normalize\_coords)}} \code{fit} measures a
per-coordinate scale $s\in\R^d$ (the coordinate standard deviation, or a
calibrated weight/activation scale) and sets
$X \mapsto \operatorname{diag}(s)X$ with the dual $\operatorname{diag}(s)^{-1}$
on queries. Equalizing coordinate ranges before a uniform rounder is exactly
the per-channel scaling of AWQ and
SmoothQuant~\cite{lin2023awq,xiao2023smoothquant}.

\prg{\code{adjust(dither)}} parameters: seed, distribution $\psi$;
$apply(x)=x+\epsilon$ for $\epsilon\sim\psi$ (e.g.\ uniform over a quantization
cell). Adding a stored pseudo-random dither decorrelates the quantization
error from the signal and makes it unbiased; the seed lets \code{reconstruct}
subtract the same $\epsilon$~\cite{roberts1962,KiracDither,feng2026rht}.

\prg{\code{adjust(center)}} \code{fit} saves $\mu=\tfrac1n\sum_i x_i$;
$apply(x)=x-\mu$; \code{score} adds the per-query constant
$\langle q,\mu\rangle$; $\code{reconstruct}(\tilde x)=\tilde x+\mu$. Centering
minimizes $\sum_i \lVert x_i-\mu\rVert^2$, shrinking the dynamic range a
downstream rounder must cover, almost for free.

\subsection{Conditioner: \code{random\_rotate}}
A \code{random\_rotate} applies $X\mapsto RX$ with a data-independent
orthogonal $R$; \code{apply\_queries} applies the same $R$ (so
$\langle Rq,Rx\rangle=\langle q,x\rangle$) and \code{reconstruct} applies
$R^T$. A random rotation spreads each vector's energy across coordinates so
that scalar or sign rounding downstream sees near-exchangeable, near-Gaussian
coordinates. The variants trade fidelity for speed.

\prg{\code{random\_rotate(full)}} \code{fit} samples a uniform (Haar)
rotation $R$; \code{apply($x$)} $=Rx$. Storing and applying $R$ requires
$O(d^2)$ space and time; the faster variants below approximate it.

\prg{\code{random\_rotate(jl)}} \code{fit} samples $R_k$, the first $k$
rows of $\sqrt{d/k}\cdot R$ (a uniform $k$-dimensional subspace);
\code{apply($x$)} $=R_k x$; $O(dk)$ time and space, lossy when $k<d$ but
inner-product preserving in
expectation~\cite{jl1984,dasgupta2003jl,achlioptas2003rp,matousek2008jl}.

\prg{\code{random\_rotate(hadamard)}} \code{fit} samples a random $\pm1$
diagonal $D$; \code{apply($x$)} $=HDx$ with $H$ the Walsh-Hadamard transform;
$O(d\log d)$ time, $O(d)$ space. This is the standard fast stand-in for a dense
random rotation~\cite{ailon2009srht,ailon2009dualbch}. 
Note that \code{random\_rotate(hadamard)} is usually applied consecutively a constant number 
of times to increase randomness. We suppress this detail in this writeup.
 
\prg{\code{random\_rotate(partial\_hadamard)}} parameter $k\le d$; the
fast Walsh-Hadamard transform evaluated at only $k$ output coordinates
($apply(x)=PHDx$ with $P$ a selection); $O(d\log k)$ time,
$O(d+k\log d)$ space~\cite{ailon2009dualbch}.

\subsection{Conditioner: \code{precondition}}
Unlike \code{random\_rotate}, a \code{precondition} map is \emph{learned}:
\code{fit} solves an optimization for a matrix $A$ (stored in the model), then
$X\mapsto AX$, $Q\mapsto (A^{\dagger})^{T}Q$, and \code{reconstruct} uses the
pseudo-inverse $\tilde A^{\dagger}$. The dual map on queries keeps dot products
consistent even when $A$ is not orthogonal.

% \prg{\code{precondition(pca)}} parameter $k\le d$; \code{fit} computes
% the SVD $X=U\Sigma V^T$ and stores $U_k$; $apply(X)=U_k^T X$; lossy when $k<d$.

\prg{\code{precondition(whitening)}} \code{fit} estimates the query
second-moment matrix $\E[qq^T]\approx U\Sigma^2U^T$ by eigendecomposition and
stores $U,\Sigma$. Then $X \mapsto \Sigma U^T X$ and
$Q \mapsto \Sigma^{-1}U^TQ$. The query map whitens the query distribution (its
transformed covariance becomes the identity) while the database map rescales
each principal direction by its query-side standard deviation, concentrating
database energy where queries actually probe; dot products are preserved
exactly since $(\Sigma^{-1}U^Tq)^T(\Sigma U^T x)=q^Tx$. This lets the
subsequent rounder allocate error in a query-aware basis. To our knowledge this
specific use of whitening for quantization is new (Table~\ref{tab:primitives},
note a).

\prg{\code{precondition(optimize\_signs)}} \code{fit} learns an
orthogonal $R$ that minimizes the reconstruction cost of the downstream
\code{sign} rounder - the ITQ rotation, found by alternating between fixing $R$
and re-binarizing (a Procrustes step);
$apply(X)=RX$~\cite{gong2013itq,ge2013opq}.

\prg{\code{precondition(optimize\_pq)}} the same skeleton, but $R$
minimizes the reconstruction cost of a downstream product quantizer (OPQ),
balancing variance across the PQ subspaces~\cite{gong2013itq,ge2013opq}.

\subsection{Rounders}
Rounders map each (transformed) vector to a codeword and, by default, emit the
residual $x-\tilde x$ downstream for further approximation; reconstruction
up-casts the codeword to full precision and \code{score} is the dot product of
the full-precision query with the reconstruction. A rounder is parameterized
by a \emph{codebook} and a \emph{rounding strategy}
(\code{closest}/\code{angular}/\code{random}/\code{adaptive}; see
\S\ref{sec:primitives}). The two families below differ only in whether the
codebook is fixed and data-agnostic (\code{cast}) or trained and stored as
the model (\code{kmeans}).

\subsection{Rounder: \code{cast}}
\code{cast} rounds to a fixed set of $2^b$ data-agnostic codewords; there is no
model, and \code{reconstruct} simply up-casts the stored code to full
precision.

\prg{\code{cast(fp16)}, \code{cast(fp8)}, \code{cast(fp4)}} round to
IEEE floating-point grids - half precision, the 8-bit \code{E4M3}/\code{E5M2}
formats, and the 4-bit \code{E2M1}/\code{E1M2} formats. \code{cast(fp16)} is
the near-lossless baseline calibrator~\cite{micikevicius2022fp8}.

\prg{\code{cast(uint,b)}, \code{cast(int,b)}} round to a uniform $b$-bit
integer lattice; unless otherwise indicated the cast is implicitly preceded by
\code{adjust(minmax)} mapping the data range onto $[0,2^b{-}1]$ (unsigned) or
$[-2^{b-1}{+}1,2^{b-1}]$ (signed)~\cite{gray1998quant}. With the
\code{adaptive} strategy the same lattice is rounded with calibration / error
feedback (GPTQ, AdaRound)~\cite{frantar2023optq,nagel2020adaround}; with the
\code{angular} strategy $x$ is rounded to the point minimizing the angle to $x$
and scored by the unbiased RaBitQ estimator (RaBitQ /
E-RaBitQ)~\cite{gao2024rabitq,gao2024erabitq}.

\prg{\code{cast(sign)}} one bit per dimension encoding the sign, $s_i=+1$
if $x_i\ge0$ else $-1$. Scoring is \emph{asymmetric}: a full-precision query is
scored against the stored signs, $\code{score}(q,s)=\langle q,s\rangle=\sum_i
q_i s_i$~\cite{gordo2014asymmetric,gao2024rabitq}.

\prg{\code{cast(hamming)}} the same one-bit-per-dimension code stored as
a bit, $b_i=1$ iff $x_i\ge0$, but scored \emph{symmetrically} between two
codes by Hamming distance,
$\code{score}=d-2\,\mathrm{popcount}(\mathrm{apply}(q)\oplus b)$ (the SimHash
estimator of the angle)~\cite{charikar2002simhash}.

\prg{\code{cast(ternary)}} dead-zone rounder to $\{-1,0,+1\}$:
$x_i\mapsto0$ when $|x_i|<\Delta$ (default $\Delta=1$ after scaling), else
$\mathrm{sign}(x_i)$~\cite{li2016twn,zhu2017ttq}.

\prg{\code{cast(nf4)}, \code{cast(beta)}} fixed 1-D codebooks matched to
an assumed coordinate distribution - the 16 normal quantiles of NF4 for
Gaussian-like weights, and a beta-distributed grid for the normalized
coordinates produced by a random rotation (the scalar code behind
EDEN/TurboQuant)~\cite{dettmers2023qlora}. A per-vector dequantization scale
$S$ is applied at reconstruct/score time
($\code{reconstruct}=S\hat y$, $\code{score}=S\langle q,\hat y\rangle$); the
choice $S=1$ (plain), $S=\langle\tilde y,\hat y\rangle/\lVert\hat y\rVert^2$
(least MSE, biased), or $S=\lVert\tilde y\rVert^2/\langle\hat y,\tilde y\rangle$
(unbiased) is what separates EDEN from EDEN-unbiased from
TurboQuant~\cite{vargaftik2022eden}.

\prg{\code{cast(MX)}} microscaling formats: small blocks share a
power-of-two scale stored as side information, with low-bit
elements~\cite{rouhani2023mx}.

\prg{\code{cast(E8)}} rounds 8-dimensional sub-vectors to the $E_8$
lattice, a fixed data-agnostic codebook that is near-optimal for the
Gaussian-like coordinates left by a random rotation; encode is
nearest-lattice-point and score is by LUT (the QuIP\#
rounder)~\cite{tseng2024quipsharp}.

\subsection{Rounder: \code{kmeans}}
\code{kmeans} learns and stores an explicit table of $2^b$ codewords as its
model; encode returns a code and \code{reconstruct}/\code{score} are table
lookups. It differs from \code{cast} only in that the codebook is trained.

\prg{\code{kmeans(lloyd)}} fit by Lloyd's algorithm over the batch;
codebook $i\mapsto\mu_i$; encode $c=\arg\min_i\lVert x-\mu_i\rVert$ (the
\code{closest} strategy). Applied per-segment this is the codebook inside
product quantization; applied to scalars it is a learned 1-D level
set~\cite{linde1980lbg,lloyd1982,Gersho1991VectorQA}.

\prg{\code{kmeans(random\_round)}} a 1-D codebook fit so that
\emph{stochastic} (unbiased) rounding to its levels has minimum variance - the
adaptive vector quantization construction~\cite{benbasat2024avq}.

\prg{\code{kmeans(angular)}} fit and encode under cosine distance
(spherical $k$-means): codewords are unit-normalized and a vector is assigned
to $\arg\max_i\langle x,\mu_i\rangle/\lVert\mu_i\rVert$, the reconstruction
scaled appropriately. Natural when only direction carries signal, e.g.\ after
\code{adjust(normalize)}.

\prg{\code{kmeans(anisotropic)}} same shape as \code{kmeans(lloyd)} for
encode/reconstruct/score, but \code{fit} minimizes a \emph{score}-weighted
(anisotropic) loss that penalizes error along the query direction more than
orthogonal to it - the ScaNN objective~\cite{guo2020scann}.

\subsection{Splitters and Routers}\label{sec:splitters}
Splitters and Routers are the multi-child primitives; both extend the
interface with arity but leave \code{fit} and \code{encode} unchanged.

A \emph{splitter} fans every vector out through \emph{all} of its children -
most often sending a different part of each vector to each child. It defines
how to decompose vectors into pieces and how to recombine the children's
reconstructions and scores:

\begin{lstlisting}
apply(model, vectors, codes) -> child_vectors[]
reconstruct(model, codes, child_recons[]) -> recons
apply_queries(model, queries) -> child_queries[]
score(model, queries, codes, child_scores[]) -> scores
\end{lstlisting}

\prg{\code{split(segment)}} \code{apply} and \code{apply\_queries} cut
each vector into disjoint, fixed-width column slices, one sub-vector per child;
\code{reconstruct} concatenates the children's reconstructions and \code{score}
sums their scores (the slices are orthogonal). This is the decomposition behind
product quantization~\cite{jegou2011pq}.

% \prg{\code{split(variable\_segments)}} parameter $k$; like
% \code{split(segment)} but with $k$ variable-width segments chosen so each
% carries roughly equal $\ell_2^2$ energy.

\prg{\code{split(outliers)}} \code{fit} identifies the outliers (whole
columns, as in LLM.int8(), or sparse per-vector entries, as in GEAR); at run
time it splits into two streams - the outliers (kept at high precision) and the
rest~\cite{dettmers2022llmint8,kang2024gear}.

\prg{\code{split(additive, m)}} $m$ \emph{full-width} sub-codebooks whose
selected entries \emph{sum} to the reconstruction (branches are not disjoint
slices); \code{fit} jointly optimizes the codebooks and the per-vector
assignment; \code{reconstruct} and \code{score} sum over branches. The greedy
sequential special case is residual
quantization~\cite{babenko2014aq,martinez2016lsq,egiazarian2024aqlm}.

\prg{\code{split(sparse\_dense)}} threshold/fraction;
\code{fit}/\code{encode} route each vector into a \emph{sparse} stream (a few
large-magnitude entries, bounded $\lVert\cdot\rVert_0$, kept at high precision)
and a \emph{dense} stream (the remainder, bounded $\lVert\cdot\rVert_\infty$,
aggressively quantized); queries pass unchanged, \code{reconstruct} adds the
two reconstructions and \code{score} sums their
scores~\cite{kim2023squeezellm,kang2024gear}.

\medskip
A \emph{router}\label{sec:routers} instead dispatches every vector through
\emph{exactly one} child branch, answering ``which branch should encode this
vector?'' - useful for recursive clustering or for separating kinds of vectors
that should be treated differently. \code{apply}, \code{reconstruct},
\code{apply\_queries}, and \code{score} carry the branch id; otherwise the
interface is unchanged:

\begin{lstlisting}
apply(model, vectors, codes) -> (child_vectors, child_ids)
reconstruct(model, codes, child_recons, child_ids) -> recons
apply_queries(model, queries, child_id) -> child_queries
score(model, queries, codes, child_scores, child_ids) -> scores
\end{lstlisting}

\prg{\code{route(norm)}} \code{fit} picks one or more norm thresholds;
each vector is dispatched on $\lVert x\rVert$ to a single branch, e.g.\ to
spend more bits on high-norm vectors. The branch id is carried through
\code{apply}/\code{reconstruct}/\code{score}.

\prg{\code{route(kmeans)}} $k$-means as a coarse quantizer: \code{fit}
learns $k$ centroids, each vector is routed to its nearest and its
\code{child\_id} emitted; downstream branches quantize each cluster (typically
the residual to the centroid) separately. This is the inverted-file
(IVF/IVFADC) coarse stage~\cite{jegou2011pq}.

% =====================================================================
\section{Chaining notation}\label{sec:chaining}
A ``\code{.}'' chains a single child below its
parent, left to right. \code{PQ = segment\_columns(width=8).kmeans(k=256)}
reads: feed every vector through \code{segment\_columns}, then feed each
resulting slice through \code{kmeans} as a vector lookup table.

\prg{Branch notation.} If a splitter or router has more than one child
of different types, it is written as a bracketed list
\code{parent.[child\_0, child\_1, ...]}; each entry is itself a (possibly
chained) sub-pipeline. For a splitter the entries are fan-out branches that
all run on every vector; for a router the entries are indexed by branch id and
each vector visits exactly one. For example,
\code{split\_norm(t=2.3).[cast\_fp16,}
\code{segment\_columns(width=8).}\allowbreak\code{kmeans(256)]}
sends vectors with $\lVert x\rVert\ge 2.3$ to an expensive 16-bit branch and
the rest to a cheap 1-bit PQ branch. A single shared sub-pipeline (as in
\code{segment\_columns}, where every slice uses the same codebook spec) is
written with the plain dot-notation chain.

% =====================================================================
\section{Quantizer details}\label{sec:quantizer-detail}
This appendix expands the catalog rows (Table~\ref{tab:catalog}) marked with
$\dagger$, where the method has additional details not captured in the pipeline notation.

\prg{EDEN / EDEN-unbiased} The two variants share the same pipeline, differing only in the per-vector scaling $S$ applied at
reconstruct/score time ($\code{reconstruct}=S\hat y$). EDEN uses $S=\langle\tilde y,\hat y\rangle/\lVert\hat y\rVert^2$ to minimize reconstruction MSE, while EDEN-unbiased uses
$S=\lVert\tilde y\rVert^2/\langle\hat y,\tilde y\rangle$ which unbiases the inner product. TurboQuant-MSE is the same
estimator with $S=1$. These three appear in Table~\ref{tab:quantizers} and the
result plots as \code{EDEN-MSE}, \code{EDEN-prod}, and \code{TurboQuant-MSE}
respectively; \code{TurboQuant-prod} spends $b-1$ bits on the same Gaussian
codebook and one bit on a QJL applied to the residual.

\prg{QuIP\#} QuIP\#'s quantizer applies incoherence
rotation (via a randomized Hadamard transform) followed by adaptive rounding of 8-dimensional blocks to the $E_8$ lattice. We omit the paper's additional step of post-quantization fine-tuning, as well as its extension to $>2$-bits.

\prg{KIVI} We implement the grouped asymmetric affine quantizer (one
scale per column group). KIVI additionally keeps a small full-precision
(fp16) window of the most recent tokens, which is out of our scope.

\prg{KVQuant} In this method, keys are scaled per channel and snapped to a \emph{shared} non-uniform codebook, given by a Lloyd-Max codebook weighted by each channel's \emph{sensitivity}, which depends on the model loss and is out of scope.

\prg{HQQ} \code{adjust(hqq)} is a \code{minmax} variant whose
offset (for a fixed scale) is optimized by a half-quadratic solver. A robust $\ell_{p<1}$ loss function prevents outliers from stretching the
offset too much.

\prg{SpinQuant} This method is equivalent to QuaRot with the fixed Hadamard replaced by an orthonormal rotation learned by Riemannian (Cayley) SGD to minimize the downstream quantization error.

\prg{Dithered-RHT} This method applies a randomized Hadamard transform, followed by a nonlinear mapping 
of the data to the unit interval, then followed by a uniform dithering step. 
The current implementation does not yet perform the nonlinear mapping method described in the paper.

% =====================================================================
\section{Quantizers: execution model}
\label{app:pseudocode}

A pipeline is a linear sequence of stages (conditioners, rounders) followed by an optional branch node (a splitter or router). To implement the four core operations of a pipeline, the harness iterates through the linear stages, then recursively delegates to the sub-pipelines at the branch. 

\prg{\code{fit}} evaluates forwards through the pipeline, partitioning at the branch. Each stage fits to the current batch of vectors and queries, then transforms the batch via \code{apply} (and \code{apply\_queries}) to produce the inputs for the subsequent stage. This enforces a strictly forward-only pipeline where each stage trains exclusively on data transformed by its predecessors, matching the exact distribution it will process during encoding. At the branch, the node itself fits first. Then, a splitter fits \emph{every} child on its respective data slice, whereas a router computes branch assignments, partitions the batch, and fits each child only on its assigned subset. Test queries are strictly withheld by the harness.

\prg{\code{encode}} evaluates forwards through the pipeline and populates a record of per-vector codes. During this forward pass, each stage appends its generated codes and side information, while propagating its \code{apply} output downstream. At a splitter, every child encodes its respective slice of the data, and the resulting fragments are concatenated. At a router, the branch identifier is explicitly stored, and children encode only their assigned subsets, yielding variable-length codes. The pipeline alone dictates memory layout and bit-packing, and as a result, the bits-per-dimension metric is computed in exactly one place.

\prg{\code{reconstruct}} evaluates backwards from the leaves of the pipeline, where the final residual corresponds directly to ``quantization loss.'' At a splitter, child outputs are recombined (e.g., by concatenating disjoint slices or summing components). At a router, child outputs are reassembled using the stored branch identifiers. These partial reconstructions then propagate upwards, where each stage applies its inverse operation: conditioners invert their transformations, and rounders add their codewords.

\prg{\code{score}} maps queries forward through the pipeline, then propagates scores backwards, without ever reconstructing the vectors. During the forward pass, the method propagates the queries via \code{apply\_queries}. The backward pass aggregates partial scores: splitters compute and combine scores from all children (e.g., summing over slices), whereas routers retrieve each vector's score from its assigned branch. These scores propagate upwards, with each stage adding its specific contribution against the query, generally without explicit reconstruction (e.g., via lookup table (LUT) access). For vector codebooks, the per-query LUT is computed exactly once and amortized across the entire database.

\prg{The harness} wraps the four calls, treating the quantizer as opaque. For each experiment, it seeds the quantizer, then calls \code{fit} (test queries withheld), \code{encode}, \code{reconstruct}, and \code{score} in order, wrapping every call with time and memory measurements. It evaluates the size of the returned code record and model to compute bits per dimension, and it computes the quality metrics of \S\ref{sec:quality} against the ground truth.

% =====================================================================
\clearpage
\section{Result plots}
\label{app:plots}

Here, we present the remaining plots for the experiment mentioned in \S\ref{sec:results}: the same four metrics (reconstruction MSE, score MSE,
recall@$10$, and attention TV distance at $\tau=0.05$) on CCNews (Figure~\ref{fig:results-ccnews}),
Yahoo (Figure~\ref{fig:results-yahoo}), COCO
(Figure~\ref{fig:results-coco}), and LAION (Figure~\ref{fig:results-laion}).
A few quantizers are omitted from the Attention TV panels for detail (noted per
figure).

\resultgrid{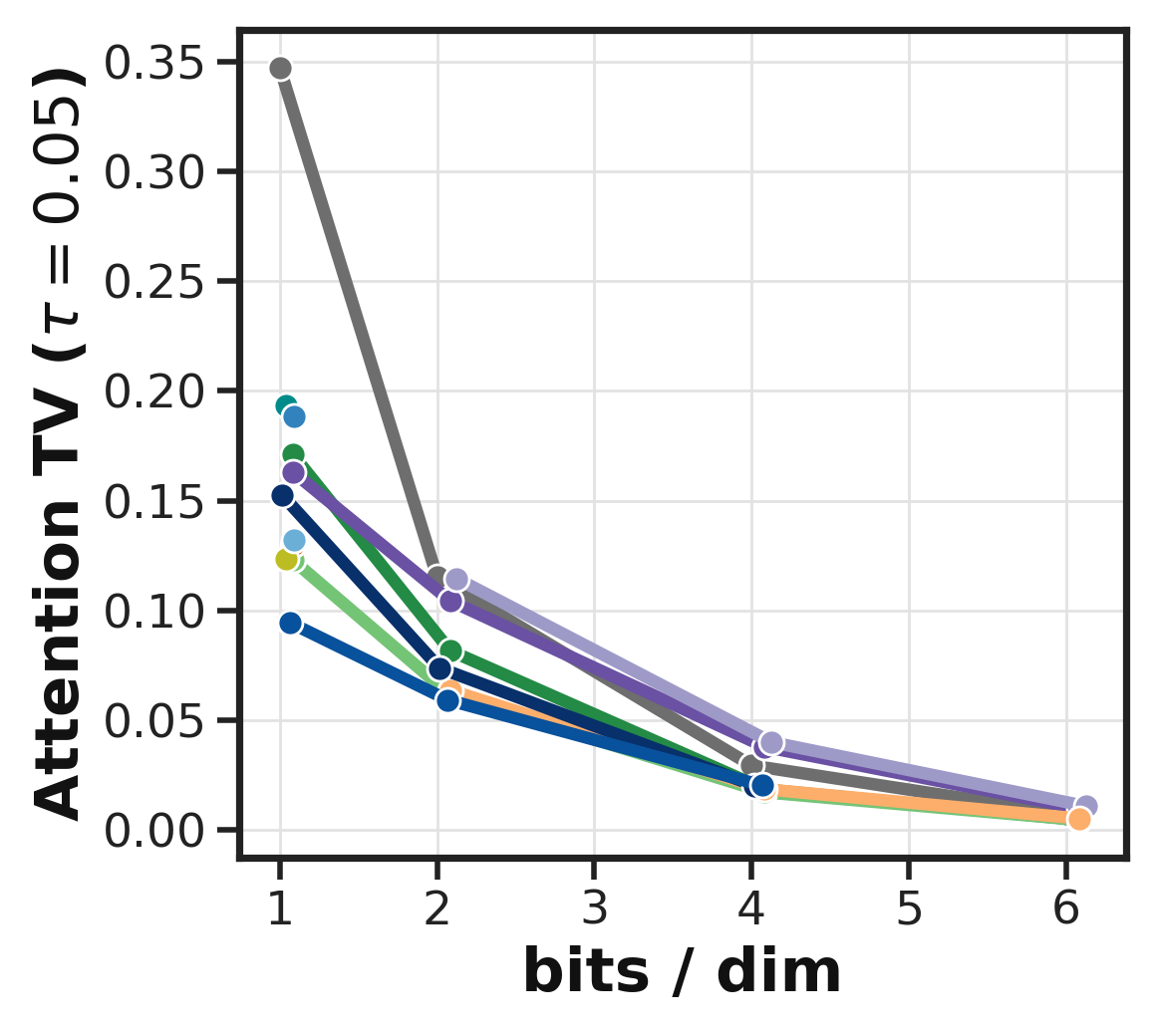}{The four headline metrics versus bits per dimension on
  CCNews. For detail, MinMax is omitted from Attention
  TV distance.}{fig:results-ccnews}

\resultgrid{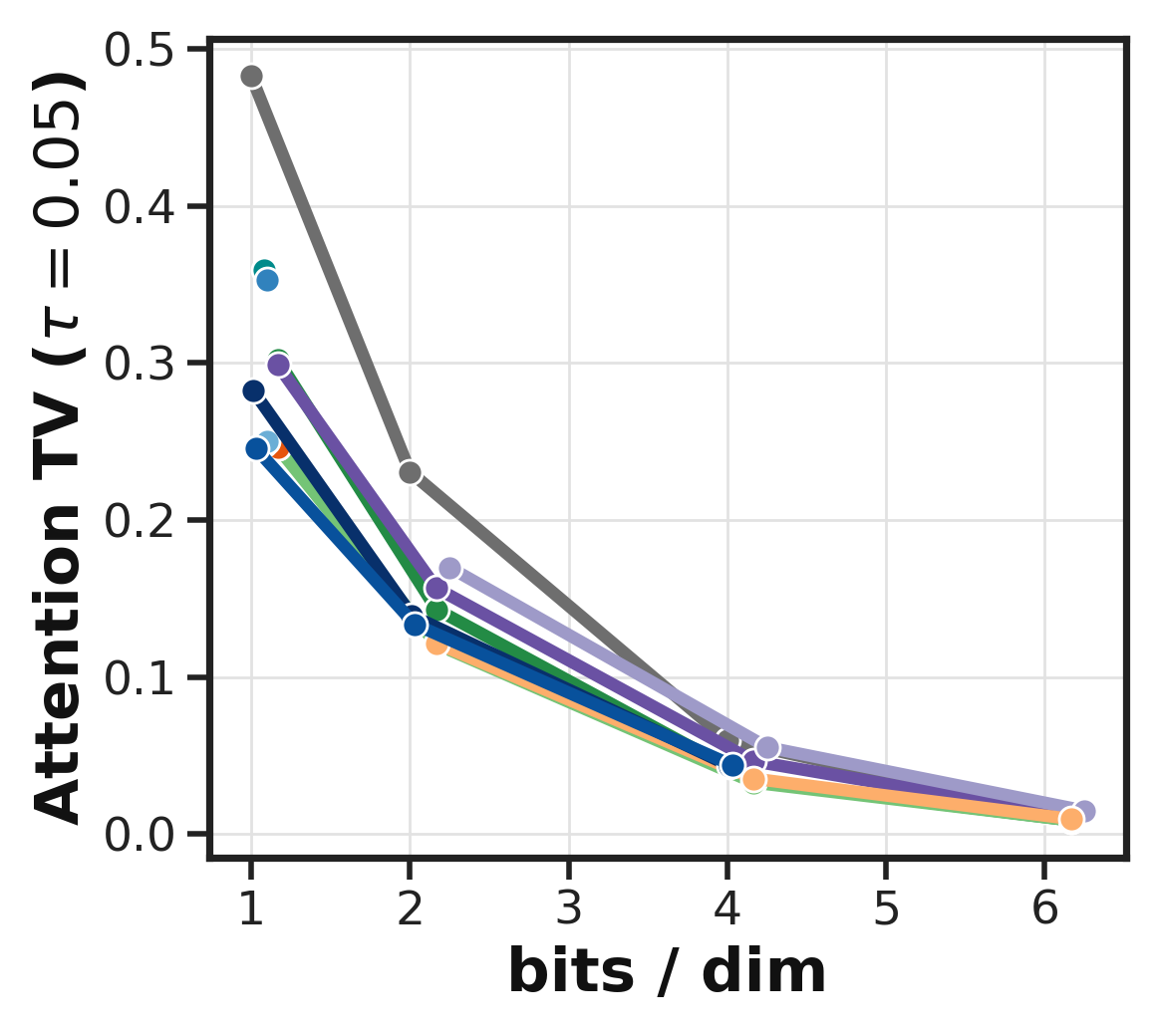}{The four headline metrics versus bits per dimension on
  Yahoo. For detail, MinMax is omitted from Attention
  TV distance.}{fig:results-yahoo}

\resultgrid{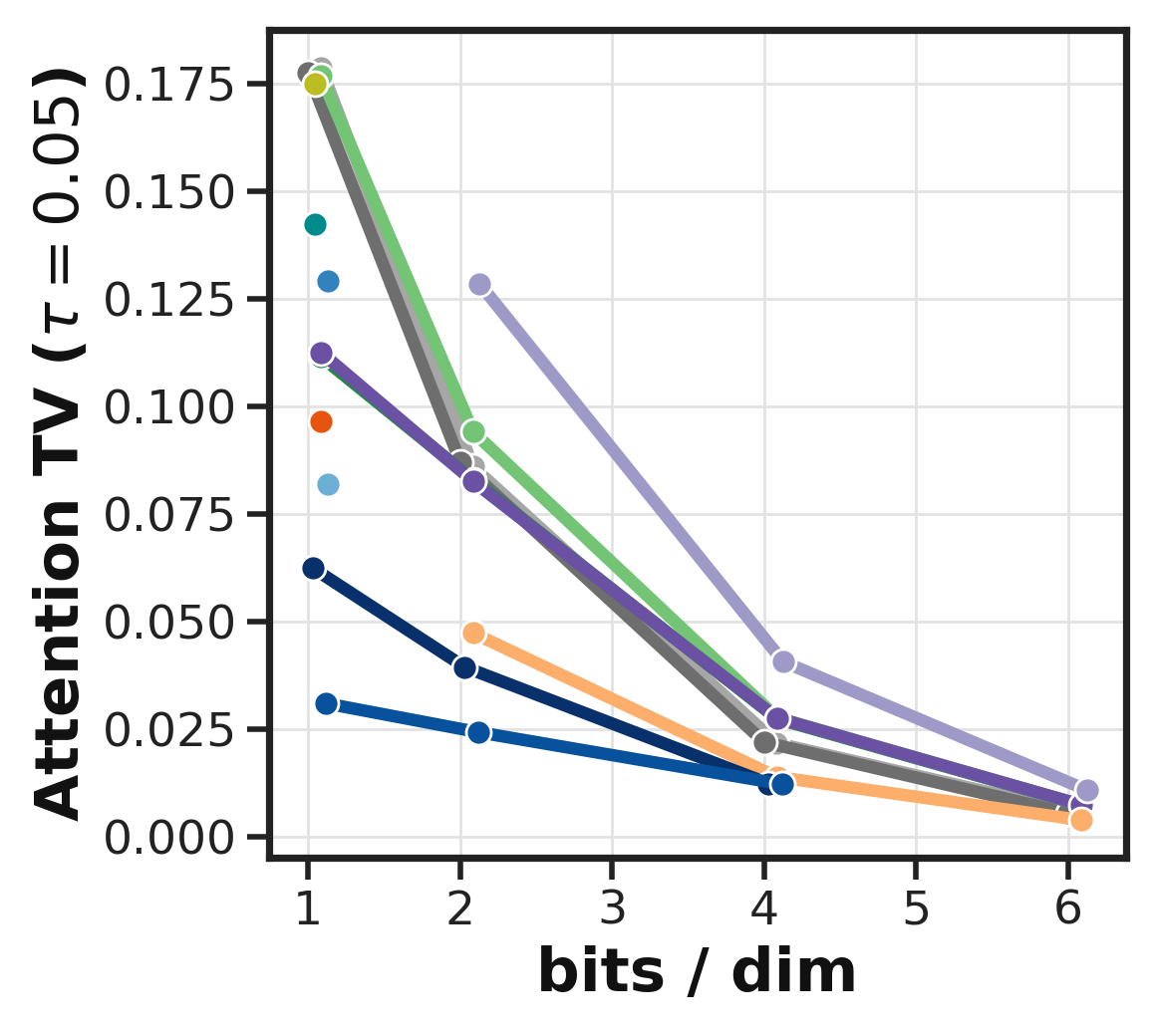}{The four headline metrics versus bits per dimension on
  COCO.}{fig:results-coco}

\resultgrid{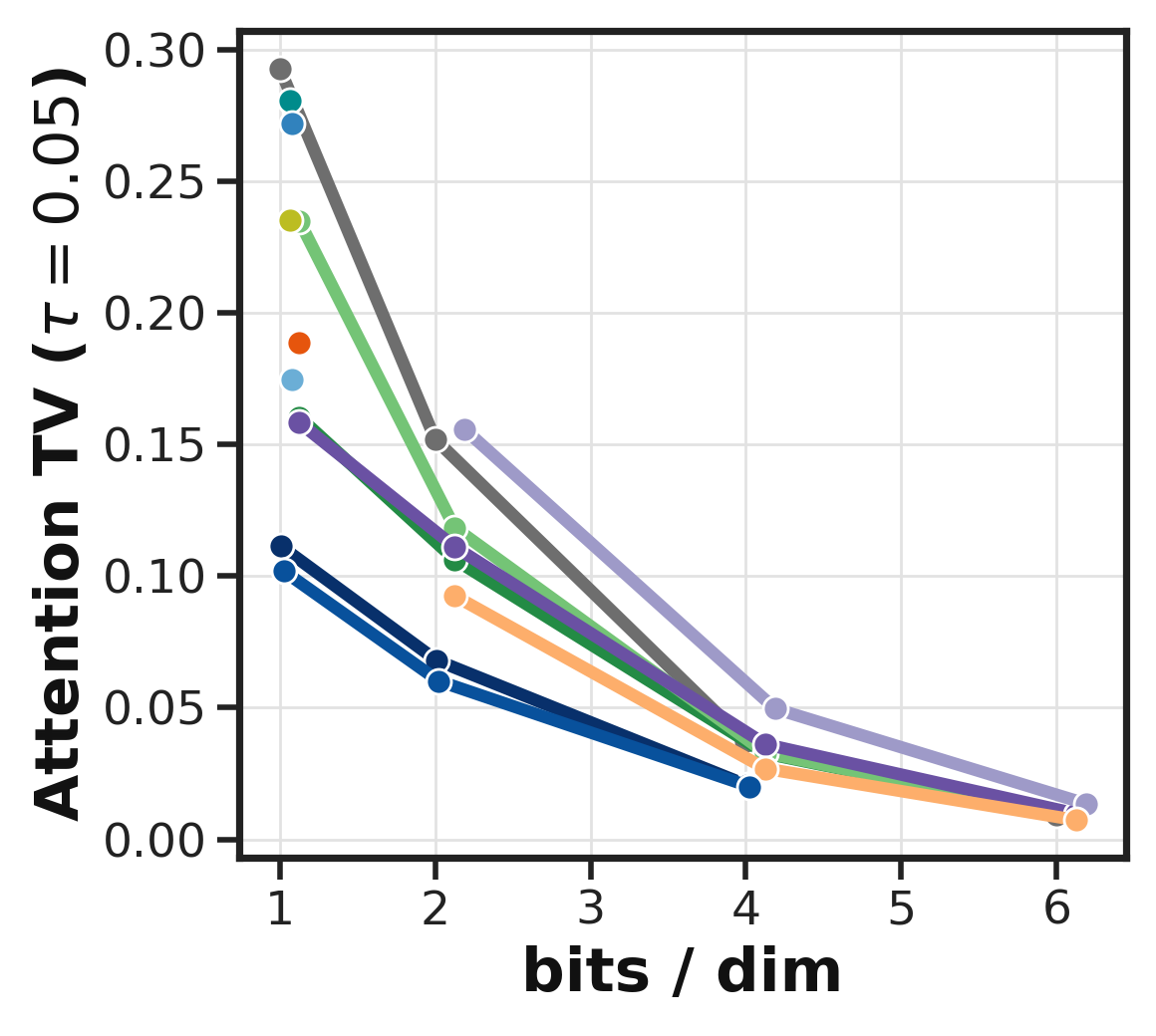}{The four headline metrics versus bits per dimension on
  LAION. For detail, MinMax is omitted from Attention
  TV distance.}{fig:results-laion}

% Flush all deferred appendix floats before acmart's end-of-document
% \balance, which otherwise silently drops them.
\clearpage

\end{document}